\documentclass[lettersize,journal]{IEEEtran}
\usepackage{amsmath,amsfonts}
\usepackage{algorithmic}
\usepackage{algorithm}
\usepackage{array}
\usepackage[caption=false,font=normalsize,labelfont=sf,textfont=sf]{subfig}
\usepackage{textcomp}
\usepackage{stfloats}
\usepackage{url}
\usepackage{verbatim}
\usepackage{graphicx}
\usepackage{cite}
\usepackage[utf8]{inputenc} 
\usepackage[T1]{fontenc}    
\usepackage{hyperref}       
\usepackage{url}            
\usepackage{booktabs}       
\usepackage{amsfonts}       
\usepackage{nicefrac}       
\usepackage{microtype}      
\usepackage{xcolor}         
\usepackage{booktabs}
\usepackage{multirow}
\usepackage{threeparttable}
\usepackage{mathtools}
\begin{document}

\title{Adaptive Hierarchical Similarity Metric Learning with Noisy Labels}

\author{Jiexi Yan, Lei Luo, Cheng Deng, ~\IEEEmembership{Senier Member,~IEEE,} Heng Huang
\thanks{Jiexi Yan and Cheng Deng are with the School of Electronic Engineering, Xidian University, Xi'an 710071, China (e-mial: jxyan1995@gmail.com, chdeng.xd@gmail.com).}
\thanks{Lei Luo is with JD Finance American Corporation, Mountain View, CA 94043 USA (luoleipitt@gmail.com).}
\thanks{Heng Huang is with the Department of Electrical and Computer Engineering, University of Pittsburgh, PA 15261 USA, and also with JD Finance American Corporation, Mountain View, CA 94043 USA (henghuanghh@gmail.com). }}

\markboth{Journal of \LaTeX\ Class Files,~Vol.~14, No.~8, August~2021}%
{Shell \MakeLowercase{\textit{et al.}}: A Sample Article Using IEEEtran.cls for IEEE Journals}


\maketitle

\begin{abstract}
Deep Metric Learning (DML) plays a critical role in various machine learning tasks.  However, most existing deep metric learning methods with binary similarity are sensitive to noisy labels, which are widely present in real-world data. Since these noisy labels often cause a severe performance degradation, it is crucial to enhance the robustness and generalization ability of DML. In this paper, we propose an Adaptive Hierarchical Similarity Metric Learning method. It considers twonoise-insensitive information, \textit{i.e.}, class-wise divergence and sample-wise consistency. Specifically, class-wise divergence can effectively excavate richer similarity information beyond binary in modeling by taking advantage of Hyperbolic metric learning, while sample-wise consistency can further improve the generalization ability of the model using contrastive augmentation. More importantly, we design an adaptive  strategy to integrate this information in a unified view. It is noteworthy that the new method can be extended to any pair-based metric loss. Extensive experimental results on benchmark datasets demonstrate that our method achieves state-of-the-art performance compared with current deep metric learning approaches.
\end{abstract}


\section{Introduction}
\IEEEPARstart{D}{eep} Deep metric learning (DML) has been actively studied recently due to its widespread applications, such as image retrieval and classification \cite{chopra2005learning,he2018triplet,zhou2017efficient}, visual tracking \cite{tao2016siamese}, person re-identification \cite{yi2014deep,yu2018hard}, and face verification \cite{hu2014discriminative,lu2017discriminative,liu2017adaptive}. The core idea of DML is to seek a reliable embedding space by virtue of nonlinear deep neural networks, where samples from the same class are encouraged to be closer than those from different classes. 

Extensive efforts  \cite{wang2017deep,oh2017deep,duan2018deep,kim2018attention,wang2019ranked,suh2019stochastic,wang2020cross,kim2020proxy,sun2020circle}  have been made to improve the performance of DML. These popular DML methods are mainly divided into two categories: loss-motivated methods and hard mining methods. For the former, the main purpose is to construct  a proper loss function to effectively characterize the real noise. To this end, a large variety of loss functions, such as Triplet loss \cite{hoffer2015deep}, Lifted Structure loss \cite{oh2016deep}, and clustering-based loss \cite{oh2017deep}, have been designed to improve the robustness of models. The other category, \textit{i.e.},  hard mining approaches,  aims  to enhance the discriminative power of the learned embedding by mining hard samples. Since training with random sampling can be overwhelmed by the redundant samples, hard sample mining has become a prevalent technique in DML \cite{oh2016deep,duan2018deep,ge2018deep}.

\begin{figure*}[!t]
	\centering
	\includegraphics[width = 1\textwidth]{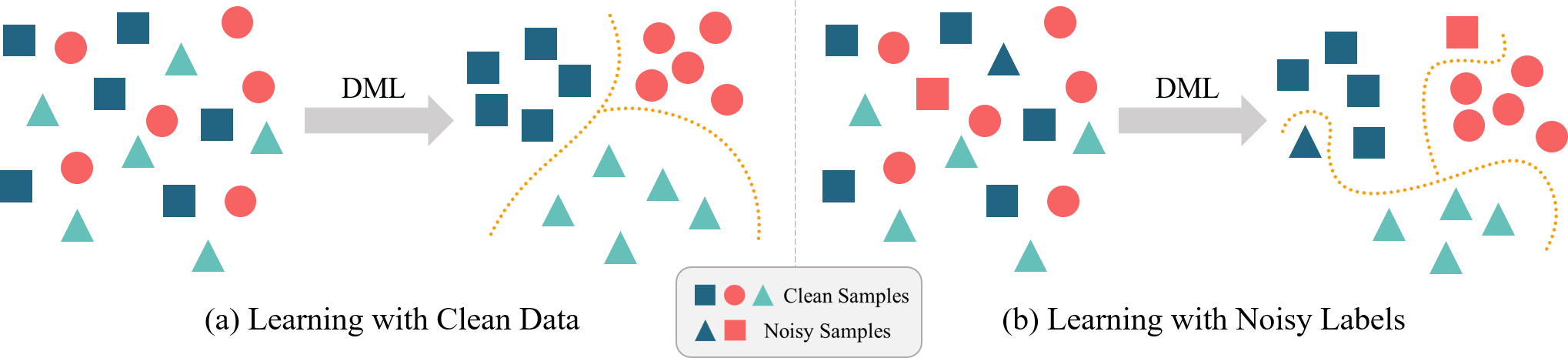}
	\caption{The simple illustration of the impact of noisy labels for traditional pair-based deep metric learning (DML). Under the influence of noisy labels, DML will pull the wrong positive samples while push wrong negative ones, which causes the learned embedding space to lose its discriminative power.}
	\label{noisy-label}
\end{figure*}

It is well-known that the superior performance of these methods heavily relies on a large-scale manual annotated dataset. However, it is expensive and inefficient to obtain such data in most real-world scenarios. To address this issue, some researchers use online key search engine \cite{li2017webvision} or crowdsourcing \cite{yu2018learning} methods to gain the desired training datasets with labels efficiently and cheaply, but noisy labels are likely to be introduced consequently. If training data contains  noisy labels, the performance of most existing DML approaches will degenerate drastically since they only excavate and  utilize the simple binary similarity information (For an anchor, the queries in the same class are identically positive while those of different classes are negative), which is very sensitive to noisy labels as shown in Figure~\ref{noisy-label}.  Therefore, it is desirable to develop robust DML algorithms for learning  an effective feature embedding under noisy labels. Considering that the mostly-used pairwise information is very sensitive to noisy labels, we need to discover noise-insensitive information for improving the robustness and generalization of DML in modeling.

\begin{figure*}[!t]
	\centering
	\includegraphics[width = 1\textwidth]{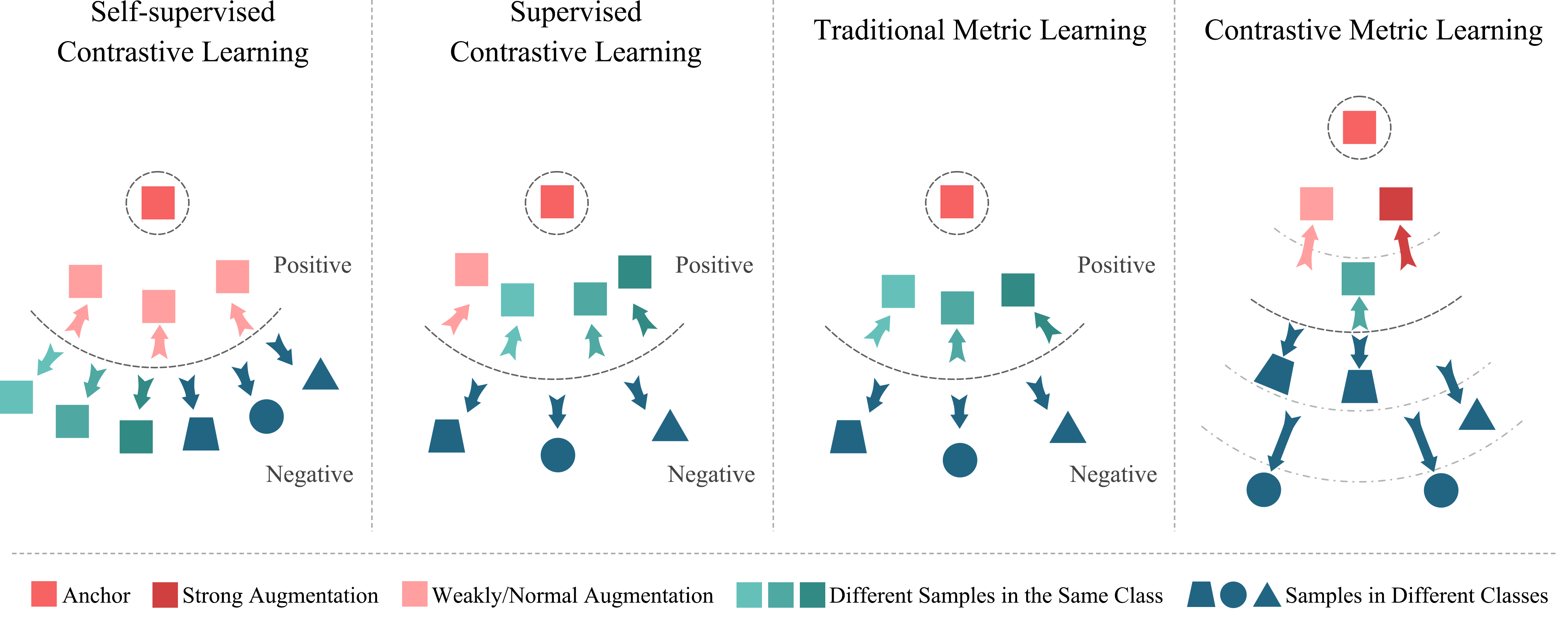}
	\caption{Comparison of similarity relations among self-supervised contrastive learning \cite{chen2020simple}, supervised contrastive learning \cite{khosla2020supervised}, traditional metric learning such as \cite{oh2016deep,wang2019multi}, and our adaptive hierarchical similarity metric learning. The first three methods all capture and use only binary similarity, \textit{i.e.}, positive and negative, which has great limitations. In contrast, our approach can mine adaptive hierarchical similarity: sample pairs will be given different distance similarities according to the dynamic hierarchical margin described in section \ref{Adaptive}. }
	\label{comparison}
\end{figure*}

In this paper, we propose a novel robust adaptive hierarchical similarity metric learning method, which integrates  class-wise divergence and sample-wise consistency into a  noise-insensitive  model to  combat noisy labels.  In real-world applications, this data often contains rich class-wise hierarchy, the traditional binary similarity can only  capture a small subset of these hierarchical relations. In view of this, we use class-wise divergence information to measure the similarity of samples from different classes. Moreover, we conduct hyperbolic geometry to sufficiently excavate the inherent information and calculate class-wise hierarchical similarity. According to this hierarchical similarity, negative pairs will be assigned dynamic margins rather than constant ones in binary similarity. Inspired by augmentation anchoring from semi-supervised learning \cite{sohn2020fixmatch}, we exploit contrastive augmentation to construct sample-wise consistency to improve the robustness and generalization ability of models. As a self-supervised strategy, augmentations are label-free positive samples, which can avoid the impact of noisy labels. Meanwhile, pulling disparate weak (\textit{e.g.}, using only crop-and-flip) and strong (\textit{e.g.}, using RandAugment \cite{cubuk2020randaugment}) augmentations can improve the generalization ability. Therefore, we introduce contrastive augmentation as nearest positive samples to construct sample-wise consistency. In the end, we propose an adaptive hierarchical similarity strategy that integrates class-wise divergence and sample-wise consistency in a unified view. It should be noted that our proposed adaptive hierarchical similarity strategy can be combined with any DML loss. As shown in Figure~\ref{comparison}, our strategy gives sample pairs hierarchical margins adaptively. Compared with existed DML methods, our proposed approach can capture more noise-insensitive similarity information, which makes the DML model more robust against noise.

The major contributions of this paper are three-fold:

\begin{itemize}
	\item \textit{Noise-Insensitive Similarity Information}: Unlike traditional DML methods with binary similarity \cite{oh2016deep,wang2019multi}, that are sensitive to noisy labels in many real-world scenarios, We propose two types of  noise-insensitive similarity information, \textit{i.e.} class-wise divergence and sample-wise consistency, to guide the DML model training robustly under the influence of noisy labels. Through  class-wise divergence and sample-wise consistency, the proposed DML model can robustly capture richer hierarchical structure of data, which cannot be discovered by the traditional binary similarity supervision. 
	\item \textit{Adaptive Hierarchical Similarity}: We design a new adaptive hierarchical similarity strategy to capture multi-level relations learned by class-wise divergence and sample-wise consistency in a unified view. Unlike existing DML methods which are only interested in binary similarity with a fixed margin, our approach can dynamically assign different margins for sample pairs to preserve the intuitive multi-level similarity in training data.
	\item \textit{State-of-the-Art Performance}: Our proposed approach achieves state-of-the-art performances on  retrieval task over three popular datasets, including CARS196, CUB-200-2011, and Standard Online Products.
\end{itemize}

The rest of this paper is organized as follows. The related work that may support and illuminate the reasons behind our work will be introduced in Section \ref{RW}. And then, we present our proposed method with detail analysis in Section \ref{Method}. Section \ref{experiment} presents our experimental results and analysis.
Finally, Section \ref{conclusion} concludes our work.

\section{Related Work}

\label{RW}

\subsection{Deep Metric Learning}
Deep metric learning methods focus on building the reliable embedding space and learning discriminative feature representations, which can be  mainly categorized into two settings: structure-learning methods and hard mining methods \cite{oh2016deep,ge2018deep,duan2018deep,duan2019deep,suh2019stochastic}. For the former, metric loss functions are built on pair-based \cite{hadsell2006dimensionality,ustinova2016learning,oh2016deep,wang2019multi,sun2020circle}, optimized by computing the pairwise similarity between samples in the embedding space,
or Proxy-based \cite{movshovitz2017no,qian2019softtriple,kim2020proxy,zhu2020fewer}, guided by comparing each sample with proxies. 

Pair-based methods rely on pairs or triplets constructed from samples in a mini-batch and  pull samples from  positive pairs together  and push samples from  negative
pairs apart from each other in the embedding space. Classical triplet loss \cite{hoffer2015deep} trains the DML model via triplets, which consist of an anchor, and corresponding positive and negative sample as follows:
\begin{equation}
	\mathcal{L}_{Triplet} = \left[ d_{an} - d_{ap} + \gamma\right]_{+},
\end{equation}
where $d_{an}$ and $d_{ap}$ denote the similarity of a negative pair and a  positive pair, respectively, and $\gamma$ is a fixed margin. Note that $\left[\cdot\right]$ is the hinge function $\left[\cdot\right]_{+} = \max(\cdot,0)$.

 Moreover, Song \textit{et al.} \cite{oh2016deep} proposed Lifted Structure Loss to utilize all the positive and negative pairs among the mini-batch as follows:
\begin{equation}
	\mathcal{L}_{Lifted} = \sum_{i = 1}^n \left[ \log \sum_{j \in \mathcal{P}_i} e^{\gamma- d_{ij}} + \log \sum_{j \in \mathcal{N}_i}e^{d_{ij}} \right]_{+},
\end{equation}
where $\mathcal{P}_i$, $\mathcal{N}_i$ represent the positive  and negative sets of the anchor $\mathbf{z}_i$, respectively, and $\gamma$ is a fixed margin.

Generally, pair-based methods can be
cast into a unified weighting formulation through General
Pair Weighting (GPW) framework \cite{wang2019multi}, \textit{i.e.}, MS loss:
\begin{equation}
	\begin{aligned}
		\mathcal{L}_{MS} = \frac{1}{n} \sum_{i = 1}^{n}  \left\lbrace \frac{1}{\varrho} \log \left[ 1 + \sum_{j \in \mathcal{P}_i} e^{-\varrho(d_{ij}-\gamma)} \right]  \right. \\
		+ \left.\frac{1}{\sigma} \log \left[ 1 + \sum_{j \in \mathcal{N}_i} e^{\sigma(d_{ij}- \gamma)} \right]\right\rbrace,
	\end{aligned}
\end{equation}
where $\mathcal{P}_i$, $\mathcal{N}_i$ represent the positive  and negative sets of anchor $\mathbf{z}_i$, respectively,  $\varrho$ and  $\sigma$ are hyper-parameters, and $\gamma$ is a fixed margin.
However, most of the mentioned
methods can only deal with binary similarity, which is sensitive to noisy labels.

\subsection{Learning with Noisy Labels}

 Learning with noisy labels is challenging in deep learning as deep neural networks (DNNs) have the high capacity to memorize the noisy labels \cite{arpit2017closer,zhang2021understanding}. Hence, it is desirable to develop robust algorithms for effectively learning with noisy labels.
 
 Extensive efforts \cite{patrini2017making,han2018co,sugiyama2018co,xia2019anchor,wei2020combating,li2020dividemix} have been made to tackle the noisy label problem for the classification task. Most approaches focus on estimating the noise transition matrix \cite{patrini2017making,xia2019anchor}, correcting the label according to model prediction \cite{ma2018dimensionality,li2020dividemix}, and training two networks on small-loss instances \cite{han2018co,wei2020combating}.
 
 Although these methods  have been proposed to handle noisy labels and achieve good performance, they cannot be directly adopted in DML since these methods mainly focus on  classification task using label information, rather than the similarity between data. Therefore, it is critical to improve the robustness and generalization of DML according to its principles of similarity information.

\subsection{Hyperbolic Geometry}
Hyperbolic geometry, also called Lobachevskian Geometry, is a non-Euclidean geometry that rejects the validity of Euclid’s fifth, the “parallel”, postulate but admits the other four Euclidean postulates.  The hyperbolic space $\mathbb{H}^d$ can be constructed using various isomorphic models such as the $d$-dimensional Poincar{\'e} ball $\mathbb{D}^d_{\tau}$ with curvature $\tau$. Intuitively, hyperbolic spaces can be thought of as continuous versions of trees, which makes it suitable for constructing hierarchical structure information. Hence, hyperbolic embeddings have become a popular technique to model real data with tree structure from network science \cite{nickel2017poincare,tifrea2018poincar,chami2019hyperbolic,mathieu2019continuous,monath2019gradient,khrulkov2020hyperbolic}.

\subsection{Self-Supervised Learning}
Self-supervised learning is widely used in unsupervised representation learning to directly  learn informative feature representations from unlabeled data \cite{caron2018deep,zhang2019aet}. Recently, contrastive learning and its variants \cite{oord2018representation,chen2020simple,khosla2020supervised,chen2020improved} develop rapidly to use augmentations as positive samples for self-supervised training supervised by the loss function as follows,
\begin{equation}
	\mathcal{L}_{i,j} = -\log \frac{\exp(d_{ij}/\nu) }{\sum^{2n}_{k= 1} \mathbb{I}_{[k \neq i]} \exp(d_{ik}/\nu)},
\end{equation}
where $\mathbb{I}_{[k \neq i]} \in \left\lbrace 0,1 \right\rbrace $ is an indicator function evaluating to 1 if $k \neq i$, and $\nu$ is the temperature parameter.
Its performance is comparable to that of supervised methods. Moreover, the augmentation idea in self-supervised learning is also adopted in some semi-supervised learning methods \cite{berthelot2019remixmatch,sohn2020fixmatch}, where  augmentation anchoring is applied to replace the consistency regularization as it shows a stronger generalization ability.

\section{Methodology}

\label{Method}

\subsection{Overview}
\textbf{Existing Problems.} 
The latest pair-based loss functions such as MS loss \cite{wang2019multi} and circle loss \cite{sun2020circle} can well capture pairwise similarity information to obtain a discriminative embedding space. 
However, in practical applications, the structure of  real-world data is more complex.  Such simple pairwise similarity cannot exactly characterize the latent relationships between the data as described in Figure~\ref{CUB}, which leads to the  pair-based methods being sensitive to noisy labels. Therefore, as shown in Figure~\ref{noisy-label}, when the training data is mixed with noisy labels, the performance of the traditional pair-based DML models will drop sharply.

\begin{figure*}[t]
	\centering
	\includegraphics[width = 1\textwidth]{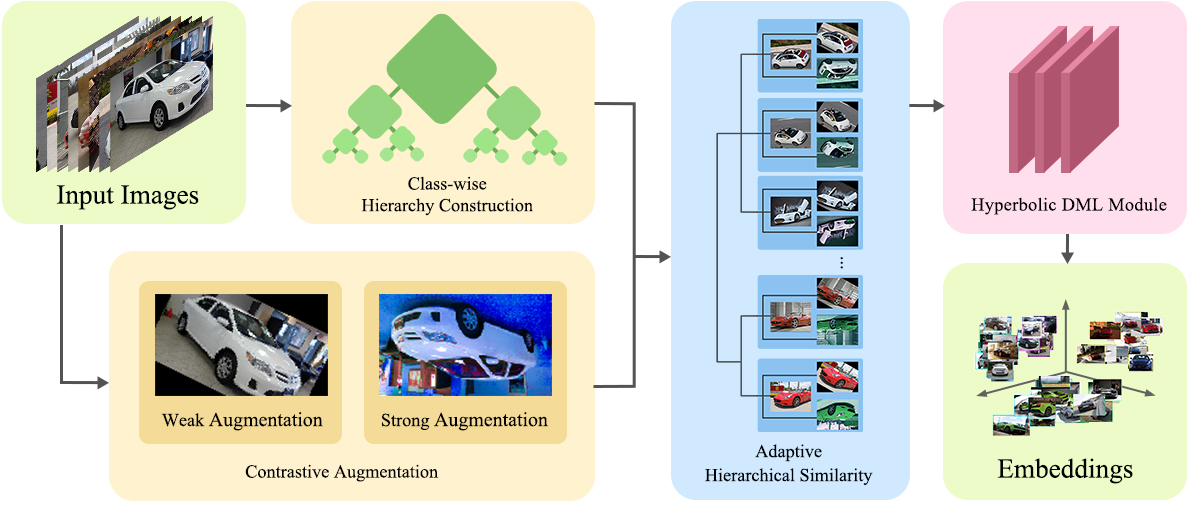}
	\caption{The pipeline of our proposed Adaptive Hierarchical Similarity Metric Learning. The model can be divided into three parts, \textit{i.e.}, class-wise hierarchy construction, contrastive augmentation, and adaptive hierarchical similarity integration. The class-wise hierarchy is calculated by intra(inter)-similarity. Combined with class-wise divergence and sample-wise consistency, the adaptive hierarchical similarity can be derived through dynamic margins $\lbrace M_a, M_p, M_n \rbrace$.}
	\label{framework}
\end{figure*}

To address the above issues, we present a novel and robust deep metric learning method to handle the data with noisy labels by using an adaptive hierarchical similarity strategy to excavate class-wise divergence and sample-wise consistency information. As shown in Figure~\ref{framework}, given the input images  $\mathcal{D} = \lbrace \mathbf{x}_1, \mathbf{x}_2, \cdots,  \mathbf{x}_n \rbrace$  with corresponding noisy labels $\mathcal{Y} = \lbrace y_1, y_2, \cdots, y_n \rbrace$, where $y_i \in \lbrace 1,2, \cdots, c \rbrace$,  we first perform contrastive augmentations $\lbrace \alpha^{-}, \alpha^{+} \rbrace$ while construct class-wise hierarchy based on the
extracted image features in hyperbolic metric space $\mathcal{Z} = \lbrace \mathbf{z}_i = f(\mathbf{x_i}| \theta) \rbrace_{i=1}^n$  through the hyperbolic metric learning module initialized by a pre-trained model. According to these information, we capture adaptive  hierarchical similarity by assigning dynamic margins $\lbrace M_a, M_p, M_n \rbrace$ to different sample pairs. Using adaptive hierarchical similarity as supervision, we can fine-tune the hyperbolic metric learning model $ f(\mathbf{x_i}| \theta)$ guided by metric loss with dynamic margins $\lbrace M_a, M_p, M_n \rbrace$. Throughout this paper, we denote the similarity between two samples $\mathbf{z}_i$ and $\mathbf{z}_j$ in the embedding space as 
\begin{equation}
	d_{ij} = \text{sim}(\mathbf{z}_i,\mathbf{z}_j) = \frac{\mathbf{z}_i^T\mathbf{z}_j}{\lVert\mathbf{z}_i\rVert\lVert\mathbf{z}_j\rVert},
\end{equation}
where $d_{ij} = $ $\lVert\cdot\rVert$ is the $\ell_2$-norm of a vector.

\subsection{Adaptive Hierarchical Similarity}
\label{Adaptive}
For better guiding the DML model with noisy labels,
we hope to discover richer noise-insensitive relation information rather
than noise-sensitive binary similarity. In view of this, we propose class-wise divergence and sample-wise consistency and integrate them into a unified adaptive hierarchical similarity strategy.

\begin{figure*}[t]
	\centering
	\includegraphics[width = 1\textwidth]{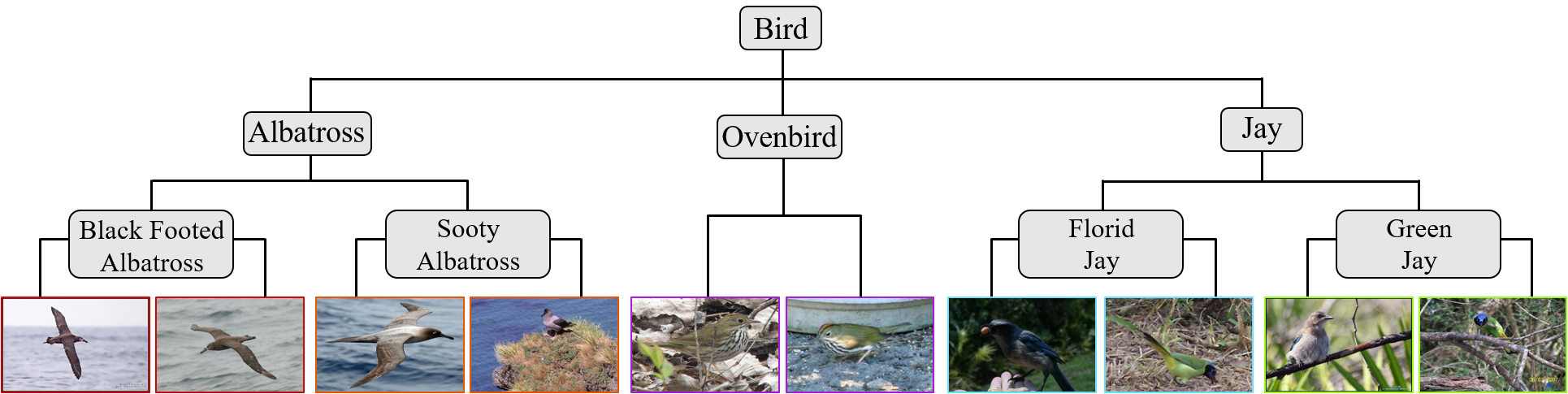}
	\caption{The natural structure in CUB dataset. Primary  class \textit{Black Footed Albatross} and \textit{Sooty Albatross} have a common secondary category label \textit{Albatross}. Meanwhile, secondary category \textit{Albatross}, \textit{Ovenbird} and \textit{Jay} all belong to the  tertiary class \textit{Bird}.} 
	\label{CUB}
\end{figure*}

\textbf{Class-wise Divergence.} In many practical applications, data usually contains  multiple classes.  For example, as shown in Figure~\ref{CUB}, in the primary similarity level, the first two birds  belong to \textit{Black Footed Albatross}, while  the third and fourth  are \textit{Sooty Albatross}. However, in the secondary similarity level, all of these four birds come from the same class \textit{Albatross}. Moreover, in the tertiary similarity level, all the ten images are \textit{Bird}. For \textit{Black Footed Albatross}, \textit{Sooty Albatross} and \textit{Ovenbird} are both negative points, but \textit{Ovenbird} should be farther than \textit{Sooty Albatross}. Such richer hierarchical similarity is more suitable to characterize the relations of samples rather than binary similarity. Meanwhile, individual sample-wise noise hardly affects the overall distribution of the categories, which makes the class-wise divergence insensitive to noisy labels. 

 To measure such a class-wise divergence, we define two distance similarity, \textit{i.e.}, average intra-similarity and average inter-similarity. The intra-similarity $S_a$ of the $a$-th class is 
\begin{equation}
	S_{aa} = \frac{2}{n_a^2 - n_a} \sum_{\mathbf{z}^a_i,\mathbf{z}^a_j \in C_a} \text{sim}( \mathbf{z}^a_i ,\mathbf{z}^a_j ),
	\label{Saa}
\end{equation}
and the inter-similarity between the  $a$-th class and $b$-th class is 
\begin{equation}
	S_{ab} = \frac{1}{n_an_b} \sum_{\mathbf{z}^a_i \in C_a,\mathbf{z}^b_j \in C_b}\text{sim}( \mathbf{z}^a_i , \mathbf{z}^b_j ),
	\label{Sab}
\end{equation}
where $\mathbf{z}^a_i$, $\mathbf{z}^b_j$ are embedding features of samples in the cluster $C_a$, $C_b$ respectively, and $n_a$, $n_b$ represent the numbers of samples in $C_a$, $C_b$, respectively. 

We map ${\mathcal{S}}_{intra} = \left\lbrace {S}_{aa} \right\rbrace_{a = 1}^c $ into $[0,0.2]$ and obtain the new set  $\hat{\mathcal{S}}_{intra} = \left\lbrace \hat{S}_{aa} \right\rbrace_{a = 1}^c$, where the maximum value of $\hat{\mathcal{S}}_{intra}$ is $0.2$ and the minimum value is $0$.
Given the positive pair in the $a$-th class, the adaptive margin with a fixed parameter $\gamma$ is defined as
\begin{equation}
	M_p = \gamma + \hat{S}_{aa}.
	\label{Mp}
\end{equation}
We also map $\tilde{\mathcal{S}}_{inter} = \left\lbrace \tilde{S}_{ab} \right\rbrace_{a \neq b} = \left\lbrace \frac{1}{S_{12}}, \frac{1}{S_{13}},\cdots,  \frac{1}{S_{1c}},\cdots, \frac{1}{S_{c-1,c}} \right\rbrace $ into $[0,0.2]$ and obtain the new set  $\hat{\mathcal{S}}_{intra} = \left\lbrace \hat{S}_{ab} \right\rbrace_{a\neq b}$, where the maximum value of $\hat{\mathcal{S}}_{inter}$ is $0.2$ and the minimum value is $0$.The dynamic margin of the negative pair between the $a$-th class and $b$-th class is represented as
\begin{equation}
	M_n =  \gamma - \hat{S}_{ab},
	\label{Mn}
\end{equation} 
where $\gamma > 0$ is a constant parameter. In addition, for better characterizing such hierarchical similarity, we also derive a hyperbolic DML paradigm, which will be introduced in section \ref{Hyperbolic}.

\textbf{Sample-wise Consistency.} Wrong labels will  mislead the DML model and degenerate its generalization ability. To alleviate this problem, we perform contrastive augmentations for each sample, weak and strong, denoted as $\alpha^{-}(\mathbf{x}_i)$ and $\alpha^{+}(\mathbf{x}_i)$, respectively. Since the augmented samples are always the positive samples regardless of the label, we can use such extra information to help the model learn more clean information without worrying about introducing noise.   Meanwhile, the stronger augmented sample $\alpha^{+}(\mathbf{x}_i)$ generates disparate
feature embedding compared with the weak one, hence maximizing sample-wise consistency of the triplet $\left\lbrace\mathbf{x}_i,\alpha^{-}(\mathbf{x}_i),\alpha^{+}(\mathbf{x}_i) \right\rbrace $ can provide better generalization performance combating noisy labels \cite{sohn2020fixmatch}. To ensure the maximum sample-wise consistency, we hope the distance between the anchor and its contrastive augmentations closer than that between the anchor and other positive queries whose labels may be wrong. 

We design the corresponding adaptive margin for the augmented pair to realize such sample-wise consistency. For the anchor in the $a$-th class, the  sample-wise consistency margin can be defined as 
\begin{equation}
	M_a = \min_{\mathbf{z}^a_i,\mathbf{z}^a_j \in C_a} \text{sim}( \mathbf{z}^a_i , \mathbf{z}^a_j).
	\label{Ma}
\end{equation}
In the experiments, we leverage a  standard crop-and-flip augmentation strategy as weak augmentation while RandAugment \cite{cubuk2020randaugment}, which is based on AutoAugment \cite{cubuk2019autoaugment}, for strong augmentation.

\textbf{Adaptive Hierarchical Similarity.} Based on the above three levels of margin (\ref{Mp}), (\ref{Mn}), and (\ref{Ma}), we can frame adaptive hierarchical similarity in a unified view. Given the anchor $\mathbf{z}_i$ in the $a$-th class, the query in its augmented set $\mathcal{A}_i $, positive set $\mathcal{P}_i$ and negative set $\mathcal{N}_i$ will be assigned dynamic margin $\left\lbrace M_a, M_p, M_n \right\rbrace $ according to the adaptive hierarchical similarity. Constrained by such multi-level similarity, we can capture the hierarchical similarity relations of data: the contrastive augmentations are the closest to the anchor as they are the noise-free positive samples, and the positive queries are closer to the anchor while different negative samples will be far away from the anchor to varying degrees. The adaptive hierarchical similarity margin can be combined with any metric loss. For example,  combined with our hierarchical margin, the MS loss \cite{wang2019multi} can be rewritten as 
\begin{equation}
	\begin{aligned}
		\mathcal{L}_{MS^*}  &=   \frac{1}{n} \sum_{i = 1}^{n}   \left\lbrace  \frac{1}{\rho}  \log   \left[  1  +   \sum_{j \in \mathcal{A}_i}   e^{ - \rho(d_{ij} - M_a)}  \right] \right.\\   
		& +  \frac{1}{\varrho}  \log  \left[  1  +   \sum_{j \in \mathcal{P}_i}   e^{ - \varrho(d_{ij} -  M_p)}  \right]\\
		&+ \left.  \frac{1}{\sigma}  \log  \left[  1  +   \sum_{j \in \mathcal{N}_i}   e^{\sigma(d_{ij} -  M_n)}  \right]\right\rbrace,
	\end{aligned}
	\label{MSn}
\end{equation}

where $d_{ij} = \text{sim} (\mathbf{z}_i , \mathbf{z}_j) $ is the similarity between the anchor $\mathbf{z}_i$ and the query $\mathbf{z}_j$, and $\rho$, $\varrho$ and $\sigma$ are fixed hyper-parameters. Algorithm \ref{Algorithm} details the iterating procedure of our proposed method.

\begin{algorithm}[t]
	\caption{Adaptive Hierarchical Similarity Metric Learning Algorithm}
	{\bf Input:}
	
	\quad \quad  Training dataset $\mathcal{D} = \left\lbrace \mathbf{x}_1, \mathbf{x}_2, \cdots, \mathbf{x}_n \right\rbrace$ with corresponding noisy labels $\mathcal{Y} = \lbrace y_1, y_2, \cdots, y_n \rbrace$;
	
	\quad \quad Hyper-parameter $\gamma$;

	\quad \quad  The number of epochs $N$.
	
	{\bf Output:} 
	
	\quad \quad Best hyperbolic metric model $f(\mathbf{x}_i|\theta)$.
	
	\begin{algorithmic}[1]
		\STATE Pre-train and initialize the parameters $\theta$.
		\FOR{ $epoch = 1,2,\cdots,N$} 
		\STATE Conduct conrastive augmentations $\lbrace \alpha^-(\mathbf{x}_i), \alpha^+(\mathbf{x}_i)  \rbrace$ for each sample following the description in subsection \ref{Adaptive}; 
		\STATE Calculate class-wise average distance matrix $\mathbf{S} = \lbrace S_{ab} \rbrace_{a,b = 1}^{c,c}$ according to Eqs.(\ref{Saa}) and (\ref{Sab});
		\STATE Produce dynamic margins $\lbrace M_a, M_p, M_n \rbrace$ with the learned adaptive hierarchical similarity using Eqs.(\ref{Ma}), (\ref{Mp}) and (\ref{Mn});
		\STATE Optimize the parameters $\theta$ using metric loss with dynamic margins, such as Eq.(\ref{MSn})  in  hyperbolic DML paradigm;
		\ENDFOR
		\STATE {\bf return} $\theta$.
	\end{algorithmic}
	\label{Algorithm}
\end{algorithm}

\subsection{Hyperbolic DML Paradigm}
\label{Hyperbolic}

Though an adaptive hierarchical similarity strategy described in section \ref{Adaptive} can well characterize multi-level similarity, traditional DML methods based on Euclidean space cannot provide meaningful geometrical representations of data, which exhibits a highly non-Euclidean latent anatomy \cite{bronstein2017geometric}.  To discover the complicated structure that implicitly exists in real data, we derive a new hyperbolic DML paradigm since the negative
curvature of the hyperbolic space is widely known to accurately
capture parent-child relationships shown in Figure~\ref{CUB} \cite{nickel2017poincare,chami2019hyperbolic}. 

Following the majority of existing works, we consider the Poincar{\'e} ball model to construct hyperbolic
space. The Poincar{\'e} ball model is defined by the manifold $\mathbb{D}_{\tau}^d = \lbrace \mathbf{x} \in \mathbb{R}^d: \tau\lVert\mathbf{x} \rVert < 1, \tau\geq 0\rbrace$, where additional hyper-parameter $\tau$ denotes    the curvature of Poincar{\'e} ball. In this model, the induced geodesic distance between any two points $\mathbf{z}_i, \mathbf{z}_j \in \mathbb{D}_{\tau}^d$  is given by the following expression \cite{nickel2017poincare}:
\begin{equation}
	d_{\mathbb{D}}(\mathbf{z}_i,\mathbf{z}_j) = \cosh^{-1}\left( 1 + 2\frac{\lVert \mathbf{z}_i - \mathbf{z}_j \rVert^2}{(1- \lVert \mathbf{z}_i\rVert^2)(1- \lVert \mathbf{z}_j\rVert^2)} \right).
\end{equation}

Towards this end, we define the  map from $\mathbb{R}^n$ to the hyperbolic manifold $\mathbb{D}^n_{\tau}$ via the $\exp$ operation in the hyperbolic space. The \textit{exponential} map $\exp^{\tau}(\mathbf{x})$ is given by:
\begin{equation}
	\exp^{\tau}(\mathbf{x}) \coloneqq \frac{\mathbf{v}}{1+2\tau\lVert\mathbf{v}\rVert^2},           
\end{equation}
where
\begin{equation}
	\mathbf{v} \coloneqq \tanh\left( \sqrt{\tau}\lVert \mathbf{x}\rVert \right)\frac{\mathbf{x}}{\sqrt{\tau}\lVert\mathbf{x}\rVert}. 
\end{equation}
In this paradigm,  we utilize such an \textit{exponential} map $\mathbf{z} = \exp^{\tau}(\mathbf{x})$ to transit the input feature vectors in Euclidean space to the Poincar{\'e} ball representations and construct the hyperbolic embedding space.

\subsection{Comparison to Other Related Methods}
Most existing pair-based metric learning methods focus on how to make full use of the pairwise binary similarity. However, the simple binary similarity cannot effectively capture the real structure information hidden in the practical data,
which limits the performance of the models, especially when the data is accompanied by label noise. Some works, such as HTL \cite{ge2018deep}, have explored how to mine hierarchical information, but HTL is only an extension of triplet loss, which cannot be compatible with other pair-based methods. In contrast, our proposed adaptive hierarchical similarity strategy can excavate richer similarity information and is easily integrated into any pair-based methods.

Although many approaches \cite{sugiyama2018co,yu2019does,wei2020combating} have been proposed to handle noisy labels, they cannot be directly adopted in DML since these methods mainly focus on  classification task using label information, rather than the similarity between data.
According to the characteristics and limitations of DML, we specially design adaptive hierarchical similarity metric learning to counteract noisy labels. In other words, this is a robust DML approach that can alleviate the impact of noisy labels, rather than a simple application of other denoise methods to DML. 

\section{Experiments}
\label{experiment}
In this section, we conduct experiments on three popular  datasets for retrieval task to demonstrate the effectiveness  and robustness of our proposed adaptive hierarchical similarity metric learning algorithm.

\subsection{Experimental Setup}

\textbf{Datasets.} We evaluate the efficacy of our proposed method on three benchmarks with simulated label noise, Cars196 \cite{krause20133d}, CUB-200-2011 \cite{wah2011caltech}, and Standard Online Products  \cite{oh2016deep}. We follow the conventional protocol \cite{oh2016deep,oh2017deep} to split them into training and test parts.
\begin{itemize}
	\item \textit{Cars196} \cite{krause20133d} (Cars) contains 16,185 car images of 196 classes. 8,054 images in the first 98 classes are used for training, and 8,131 images in the remaining 98 classes are used for testing. 
	
	\item \textit{CUB-200-2011} \cite{wah2011caltech} (CUB) includes 11,788 images of 200 bird species. We use the first 100 classes (5,864 images) for training and the other 100 classes (5,924 images) for testing. 
	
	\item \textit{Standard Online Products} \cite{oh2016deep} (SOP) is composed of 120,503 images of 22,634 products from eBay.com. We use the first 11,318 products with 59,551 images and the other 11,316 products with 60,502 images for training and testing, respectively.
	
\end{itemize}

\begin{table}[]
		\caption{Descriptions of the experimental datasets.}
		\renewcommand{\arraystretch}{1.5}
		\centering
	\begin{tabular}{c|cc|cc}
		\hline
		\multirow{2}{*}{Dataset} & \multicolumn{2}{c|}{Training} & \multicolumn{2}{c}{Testing} \\ \cline{2-3}  \cline{4-5} 
		& \# Samples     & \# Classes    & \# Samples    & \# Classes    \\ \hline \hline 
		Cars                 &     8,054          &     98         &       8,131       &      98        \\
		CUB             &     5,864          &      100        &      5,924        &        100      \\
		SOP &       59,551        &       11,318       &     60,502         &        11,316      \\ \hline
	\end{tabular}
\label{data}
\end{table}

\begin{table*}[]
	\caption{Comparison with the state-of-the-art methods on clean datasets. The performances of retrieval are measured by Recall@$K$ (\%). “–” means that the result is not available from the original paper. Triplet$^{*}$, LiftedStruct$^{*}$ and MS$^{*}$ represent our method combined with Triplet, LiftedStruct and, MS respectively. }
	\resizebox{\textwidth}{!}{%
		\renewcommand{\arraystretch}{1.5}
		\centering
		\begin{tabular}{c|cccc|cccc|ccc}
			\hline
			\multirow{2}{*}{Method} & \multicolumn{4}{c|}{Cars196} & \multicolumn{4}{c|}{CUB-200-2011} & \multicolumn{3}{c}{Standard Online Products} \\ \cline{2-12} 
			& R@1  & R@2  & R@4 & R@8  & R@1  & R@2  & R@4  & R@8   & R@1  & R@10  & R@100  \\ \hline
			\hline
			ProxyNCA \cite{movshovitz2017no}&   73.2   &   82.4   &    86.4  &   88.7   &  49.2    &  61.9    &   67.9   &   72.4   &   73.7   &       -    &    -\\
			HDC \cite{yuan2017hard}	&   73.7   &   83.2   &   89.5   &   93.8   &   53.6   &   65.7   &   77.0   &   85.6   &   69.5   &    84.4       &  92.8  \\
			HTL \cite{ge2018deep}&   81.4   &   88.0   &  92.7    &   95.7   &   57.1   &   68.8   &  78.7    & 86.5     &   74.8   &   88.3        &  94.8  \\
			ABE \cite{kim2018attention}	&  85.2    &   90.5   &   94.0   &   96.1  &   60.6   &   71.5   &  79.8    &   87.4   &   76.3   &    88.4       &   94.8  \\
			SoftTriple \cite{qian2019softtriple}	&  84.5    &   90.7   &   94.5   &   96.9   &   65.4   &  76.4    &   84.5   &   90.4  &     78.3  &      90.3    & 95.9  \\
			ProxyGML \cite{zhu2020fewer}	&   85.5   &  91.8    &    95.3  &   -   &   66.6   &   77.6   &   86.4   &   -   &  78.0    &      90.6     &   96.2  \\
			CircleLoss \cite{sun2020circle}	&  83.4    &   89.8   &   94.1   &  96.5    &   66.7   &  77.4    & 86.2      &   91.2   &   78.3   &   90.5   &    96.1      \\ \hline
			Triplet	\cite{schroff2015facenet}&   46.2   &   58.4   &  69.5    &  78.4    &   55.1   &   67.7   &   77.8   &    85.8  &       58.9  &   75.4   &    87.9    \\
			Triplet$^{*}$&	\textbf{47.8}		&   \textbf{59.6}   &  \textbf{70.2}    &   \textbf{80.1}   &  \textbf{56.8}    &   \textbf{68.6}   &   \textbf{78.9}   &   \textbf{87.0}   &    \textbf{59.6} &   \textbf{76.1}   &    \textbf{88.3}        \\\hline
			LiftedStruct \cite{oh2016deep} 	&   80.5   &   87.2   &   91.9   & 95.1 &   63.6   &  74.8    & 83.5     &  89.8    &   69.9   &    83.4      &  92.4  \\
			LiftedStruct$^{*}$ 		&   \textbf{81.6}   &  \textbf{87.9}    &   \textbf{92.4}   &   \textbf{95.3}   &    \textbf{64.4}    &   \textbf{75.1}   &   \textbf{83.9}   &   \textbf{90.4}     &    \textbf{71.2}      &   \textbf{83.9}   &   \textbf{92.7} \\ \hline
			MS \cite{wang2019multi}&  84.1    &   90.4   &   94.0   &   96.5   &   65.7   &   77.0   &    86.3  &  91.2    &  78.2     &   90.5   &   96.0 \\
			MS$^{*}$ &      \textbf{85.4}   &  \textbf{91.2}    &  \textbf{94.5}   &  \textbf{96.7}    &   \textbf{66.8}   &   \textbf{77.5}   &   \textbf{86.4}   &   \textbf{91.3}   &    \textbf{78.5}  &    \textbf{90.7}      &   \textbf{96.2}  \\
		\end{tabular}
	}
	\label{clean}
\end{table*}

The details of the datasets are summarized in Table \ref{data}. These datasets are initially clean. To verify the robustness of the model, we also inject  random label noise by randomly flipping the labels of a given portion of data samples, while keeping the test sets clean.

\textbf{Evaluation Metric.} Following the standard protocol \cite{sohn2016improved,oh2016deep}, our method and compared approaches are evaluated on the image retrieval task. For the retrieval task, we use the standard performance metric Reall@$K$, \textit{i.e.}, computing the percentage of testing samples which have at least one example from the same category in $K$ nearest neighbors.

\textbf{Implementation Details.} Our method is implemented in PyTorch \cite{paszke2019pytorch} with an NVIDIA TITAN XP GPU of 12GB memory. Following  the standard data pre-processing paradigm, we normalized the input images into $256 \times 256$ at first and then cropped them to $224 \times 224$. We employ Inception \cite{ioffe2015batch}
pre-trained on the ImageNet  dataset  \cite{russakovsky2015imagenet} as our backbone feature embedding network with the embedding dimension as $512$. Moreover, on top of the network following
the global pooling layer, a fully-connected layer was added
with random initialization. Adam optimizer \cite{kingma2014adam} is used in all experiments and the weigh decay is set $1e^{-5}$. In the all experiments, we set $\gamma = 0.5$.

\subsection{Performance Comparisons with State-of-the-Art Methods}
We implement our adaptive hierarchical similarity metric learning framework with three aforementioned baseline pair-based loss functions, including Triplet loss (Triplet) \cite{schroff2015facenet} and Lifted-Structure loss (LiftedStruct) \cite{oh2016deep} and Multi-Similarity loss (MS) \cite{wang2019multi}. We compare the performance of baseline methods before and after applying our Adaptive Hierarchical Similarity strategy to demonstrate the effectiveness of our proposed approach. Meanwhile, we compare the performance of our proposed methods with other state-of-the-art methods, including pair-based methods: HDC \cite{yuan2017hard}, HTL \cite{ge2018deep}, Circle loss \cite{sun2020circle}; proxy-based methods: ProxyNCA and ProxyGML \cite{zhu2020fewer}; and other recent methods: ABE \cite{kim2018attention} and SoftTriple \cite{qian2019softtriple}. 

\textbf{Comparison on clean datasets.} The results for the original  Cars196, CUB-200-2011,  and Stanford Online Products datasets are summarized in Table \ref{clean}.  Note that bold numbers represent the improved results of original metric learning approaches by our proposed method. We observe that the proposed adaptive hierarchical metric learning boosts the performance of original metric learning approaches on all the benchmark datasets. Moreover, combined with Multi-Similarity loss \cite{wang2019multi}, our method can also achieve comparable results, especially on CUB-200-2011, our method has the best Recall@$1$ value $66.8\%$. This demonstrates the effectiveness of our method. Due to the adaptive hierarchical similarity, our approach can learn more intrinsic semantic information for training. 

\textbf{Comparison on datasets with $30\%$ noisy labels.} Table \ref{noise-results} reports the retrieval results of our proposed adaptive hierarchical similarity metric learning compared with baseline methods on the simulated Cars196, CUB-200-2011,  and Stanford Online Product datasets with $30\%$ noisy labels, respectively. In the tables, bold numbers represent that our approach improves the results of the original metric learning algorithms.  The performance of baseline methods decreases significantly over the training data with $30\%$ noisy labels, which shows that those original baseline methods are sensitive to noisy labels. However, in this difficult case, our proposed method can also outperform the baseline approaches by a large margin, \textit{e.g.} improving the Recall@$1$ from $62.0\%$ to $65.3\%$ over CUB-200-2011, better than the baseline method by $5.3\%$. Overall, the result shows that our proposed adaptive hierarchical similarity strategy alleviates the impact of noisy labels due to the extra noise-insensitive information, \textit{i.e.}, class-wise divergence, and sample-wise consistency. 

\begin{figure*}[t]
	\centering
	\includegraphics[width = 1\textwidth]{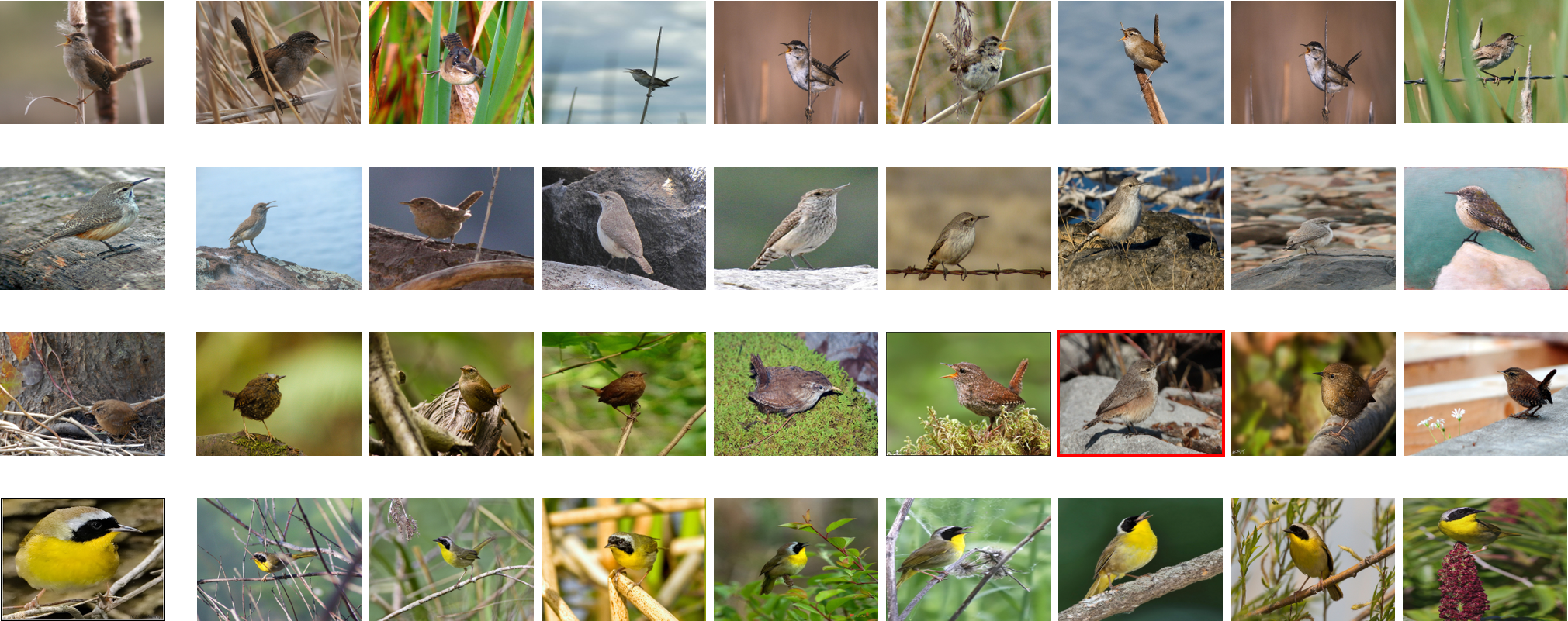}
	\caption{The top 8 images retrieved based on the  embedding features learned by our robust DML model on the  CUB-200-2011 \cite{wah2011caltech}. The red borders indicate that the retrieved images do not belong to the correct class.}
	\label{visualization}
\end{figure*}

\textbf{Visualization of the retrieval results.} The images retrieved using the embedding features learned by  our model are shown in Figure \ref{visualization}.
Red borders indicate wrongly returned images. From these results, we can observe that our proposed method can effectively learn good feature embedding under the influence of noisy labels.

\begin{table*}[]
	\caption{Comparison with the state-of-the-art methods on simulated datasets with $30\%$ noisy labels. The performances of retrieval aremeasured by Recall@$K$ (\%). Triplet$^{*}$, LiftedStruct$^{*}$ and MS$^{*}$ represent our method combined with Triplet, LiftedStruct, and MS respectively.}
	\resizebox{\textwidth}{!}{%
		\renewcommand{\arraystretch}{1.6}
		\centering
		\begin{tabular}{c|cccc|cccc|ccc}
			\hline
			\multirow{2}{*}{Method} & \multicolumn{4}{c|}{Cars196} & \multicolumn{4}{c|}{CUB-200-2011} & \multicolumn{3}{c}{Standard Online Products} \\ \cline{2-12}  
			& R@1  & R@2  & R@4 & R@8  & R@1  & R@2  & R@4  & R@8 & R@1  & R@10  & R@100  \\ \hline \hline
			Triplet	\cite{schroff2015facenet}&  44.3&   57.0   &   69.0   &   79.1   &   54.3   &   67.1   &  77.4    &   85.6   &    51.7        &  69.2    &  84.1     \\
			Triplet$^{*}$&  \textbf{46.1}    &   \textbf{58.2}   &   \textbf{69.6}   & \textbf{79.3}    &  \textbf{55.5}    &   \textbf{68.1}   &   \textbf{78.2}   &   \textbf{85.9}   &   \textbf{52.9}        &  \textbf{70.1}   &  \textbf{84.6}  \\\hline
			LiftedStruct \cite{oh2016deep}&   77.1   &  85.3    &   91.6   &  94.8    &   61.6   &  73.0    &  82.1    &   89.1   &     67.9      &   82.0   &  91.5  \\
			LiftedStruct$^{*}$ 		&     \textbf{79.2}   &  \textbf{87.1}    &   \textbf{92.0}   &   \textbf{95.0}    &  \textbf{64.3}    &   \textbf{75.5}   &   \textbf{83.6}   &   \textbf{90.1}   &     \textbf{69.1}     &    \textbf{83.0}  &   \textbf{92.1}       \\ \hline
			MS 	\cite{wang2019multi}&  79.5    &  86.7    &  91.7    &   95.1   & 62.0 &    73.8  &   82.5   &   89.6     &    72.0   &   85.7     &   94.1 \\
			MS$^{*}$ &    \textbf{82.4}  &  \textbf{89.5}    &  \textbf{93.8}  & \textbf{95.9}   &  \textbf{65.3} &    \textbf{76.1}  &   \textbf{84.7}   &   \textbf{90.7}   &  \textbf{73.6}  &  \textbf{86.9}  &   \textbf{94.8} \\
		\end{tabular}
	}
	\label{noise-results}
\end{table*}

\subsection{Ablation Study}
In this subsection, we conduct ablation studies to perform in-depth analysis of our proposed method and  discuss the impact of its different components to model performance on the datasets with different noise ratio. Here, we use Lifted Structure loss \cite{oh2016deep} as baseline, since it is a classical pair-based DML method, whose effectiveness have been widely recognized.

\textbf{Performance contributions of different components.} We ablate our proposed Adaptive Hierarchical Similarity Metric Learning to evaluate the effectiveness of different components. Our proposed
method contains two major components: class-wise divergence implemented by Eqs. (\ref{Mp}) and (\ref{Mn}), and sample-wise consistency implemented by Eq. (\ref{Ma}). In
order to evaluate the effectiveness and robustness of different components of the proposed method, we conduct  DML task on the CUB-200-2011 \cite{wah2011caltech} with $\lbrace 30\%, 50\%, 70\% \rbrace$ noisy labels using different method configurations: (1) Baseline with class-wise divergence; (2) Baseline with sample-wise consistency; (3) Baseline + full of our method as the baseline. 

The experimental results are summarized in Table \ref{ablation}. It is seen that both the class-wise divergence and the sample-wise consistency significantly improve the performance of baseline combating noisy labels.

Moreover, we evaluate the effectiveness of the hyperbolic geometry by conducting retrieval task on the  CUB-200-2011 \cite{wah2011caltech} with $ 30\%$ noisy labels. As shown in Table \ref{HG}, hyperbolic geometry (HG) can  boost the performance of our method, which demonstrates the effectiveness of hyperbolic geometry. 

\textbf{Impact of different noise ratios.} In order to further analyze the robustness and generalization ability of the proposed method, we compare our approach with the baseline on unclean CUB-200-2011 \cite{wah2011caltech} with different noise ratios in the range of $\lbrace 30\%, 50\%, 70\% \rbrace$. The results for Recall@$1, 2, 4, 8$ are presented in Table \ref{ablation}. It is obvious that our proposed method can improve the performance of the baseline under all noise ratio cases. Even in the hardest $70\%$ noisy labels case,  the Recall@$1$ of our method is still higher than the baseline by $7.2\%$. This further illustrates the effectiveness and robustness of our approach.

\begin{table*}[!t]
	\caption{The performance of different components from our method on the CUB-200-2011 \cite{wah2011caltech} with different noise ratio.}
	\resizebox{\textwidth}{!}{%
		\renewcommand{\arraystretch}{1.7}
		\centering
		\begin{tabular}{c|cccc|cccc|cccc}
			\hline
			\multirow{2}{*}{Method} & \multicolumn{4}{c|}{30\%} & \multicolumn{4}{c|}{50\%} & \multicolumn{4}{c}{70\%} \\ \cline{2-13} 
			& R@1  & R@2  & R@4 & R@8  & R@1  & R@2  & R@4  & R@8  & R@1  & R@2  & R@4  & R@8 \\ \hline
			\hline
			Baseline \cite{oh2016deep}	&  61.6    &  73.0   &   82.1   &   89.1   &  56.5    &  69.7   &  79.7    &   87.5   &  50.2    &  63.8    &  75.9    &  85.7  \\
			+ Class-wise Divergence		& 62.7     &   74.4   &   83.4   &   89.7   &   57.9   &   70.6   &  81.2    &   88.4   &  51.3    &   64.9   &  76.3    &  85.9  \\ 
			+ Sample-wise Consistency&  63.1    &   74.4   &   83.7   &   89.9       &   58.4   &   70.8   &  81.0    &  88.6   &    52.3  &   66.1   &  77.8 & 86.8  \\\hline
			Ours		&    \textbf{64.3 }    &   \textbf{75.5}   &   \textbf{83.6}   &   \textbf{90.1 }      &  \textbf{59.7}    &   \textbf{71.3}   &  \textbf{81.4}   &   \textbf{88.9}  &  \textbf{53.8}    &   \textbf{66.6}   &   \textbf{78.6}   &   \textbf{87.1} \\
		\end{tabular}
	}
	\label{ablation}
\end{table*}

\begin{table}[]
	\caption{The performance of  our method with (without) hyperbolic geometry on the CUB-200-2011 \cite{wah2011caltech} with $ 30\%$ noisy labels.}
	\renewcommand{\arraystretch}{1.7}
	\centering
	\begin{tabular}{c|cccc}
		\hline
		Method      & Recall@1 & Recall@2 & Recall@4 & Recall@8 \\ \hline \hline
		Baseline  \cite{oh2016deep}    &    61.6      &      73.0    &     82.1     &     89.1     \\
		Ours w/o HG &     63.7     &    75.1     &    83.3      &     89.9     \\ \hline
		Ours        &      \textbf{64.3 }    &   \textbf{75.5}   &   \textbf{83.6}   &   \textbf{90.1 }      
	\end{tabular}
	\label{HG}
\end{table}

\section{Conclusion}
\label{conclusion}
In this paper, we proposed a novel Adaptive Hierarchical Similarity Metric Learning approach combating noisy labels. Considering the limitation of most frequently-adopted binary similarity, which was mostly used in traditional pair-based deep metric learning, we design an adaptive hierarchical similarity strategy to mine richer multi-level similarity that implicitly exists in data, which can effectively alleviate the influence of noisy labels. The adaptive hierarchical similarity is composed of class-wise divergence and sample-wise consistency. Class-wise divergence can learn more noise-insensitive hierarchical information to  boost the performance of the model, while sample-wise consistency utilizes the self-supervised contrastive augmentations to 
enhance robustness and generalization ability of the metric model. Moreover, we also derive a hyperbolic deep metric learning paradigm for better characterizing multi-level similarity. The experimental results on Cars196, CUB-200-2011,  and Stanford Online Products benchmarks demonstrate the superiority of our proposed method over several state-of-the-art DMLs.

\bibliographystyle{IEEEtran}
\bibliography{Adaptive_Hierarchical_Similarity_Metric_Learning_with_Noisy_Labels}

\begin{IEEEbiography}[{\includegraphics[width=1in,height=1.25in,clip,keepaspectratio]{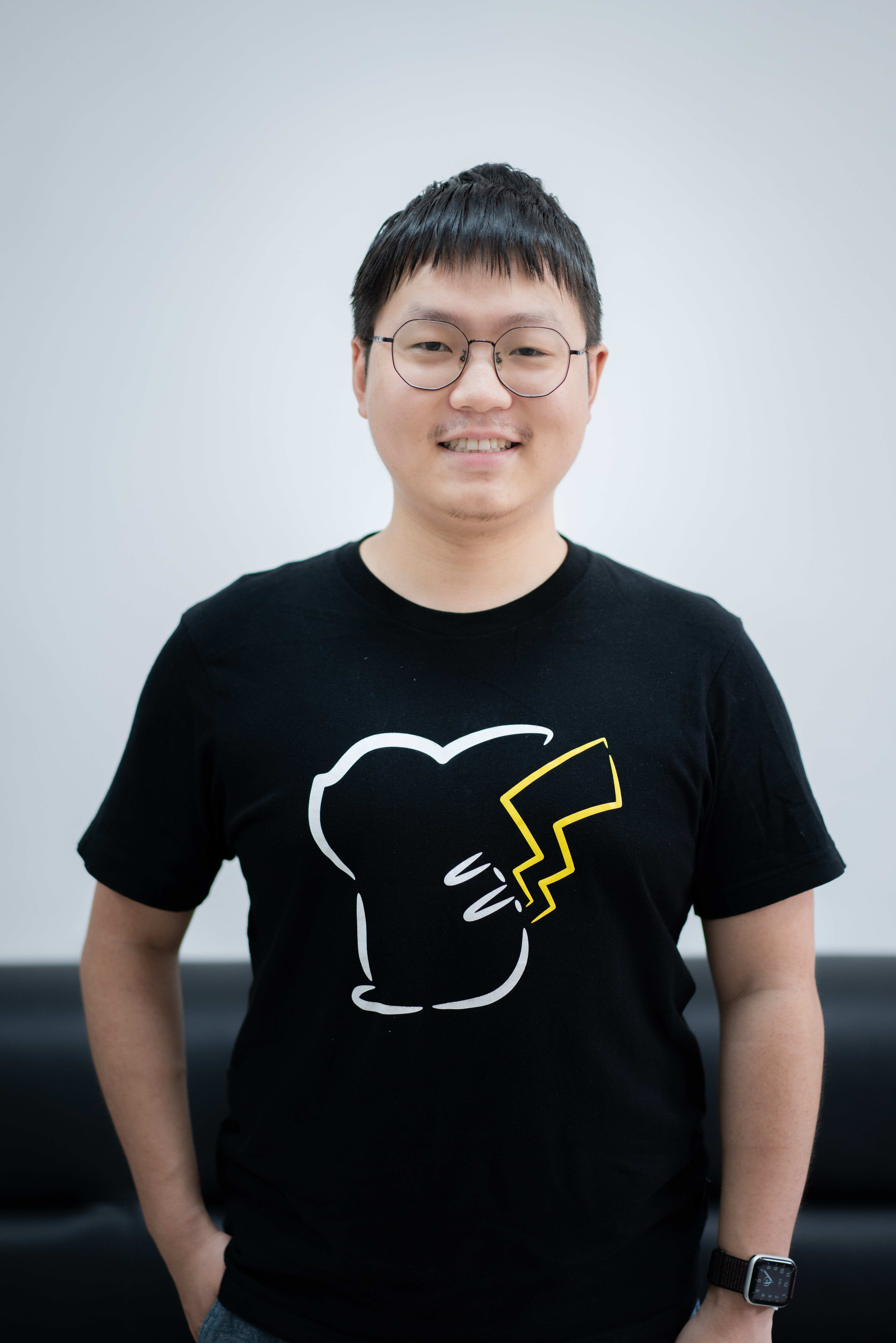}}]{Jiexi Yan}
	received the B.E. degree in electronic and information science and technology from Xidian University, Xi’an, China, in 2017, where he is currently
	pursuing the Ph.D. degree with the School of Electronic Engineering.
	His current research interests include machine
	learning and computer vision.
\end{IEEEbiography}

\vspace{11pt}

\begin{IEEEbiography}[{\includegraphics[width=1in,height=1.25in,clip,keepaspectratio]{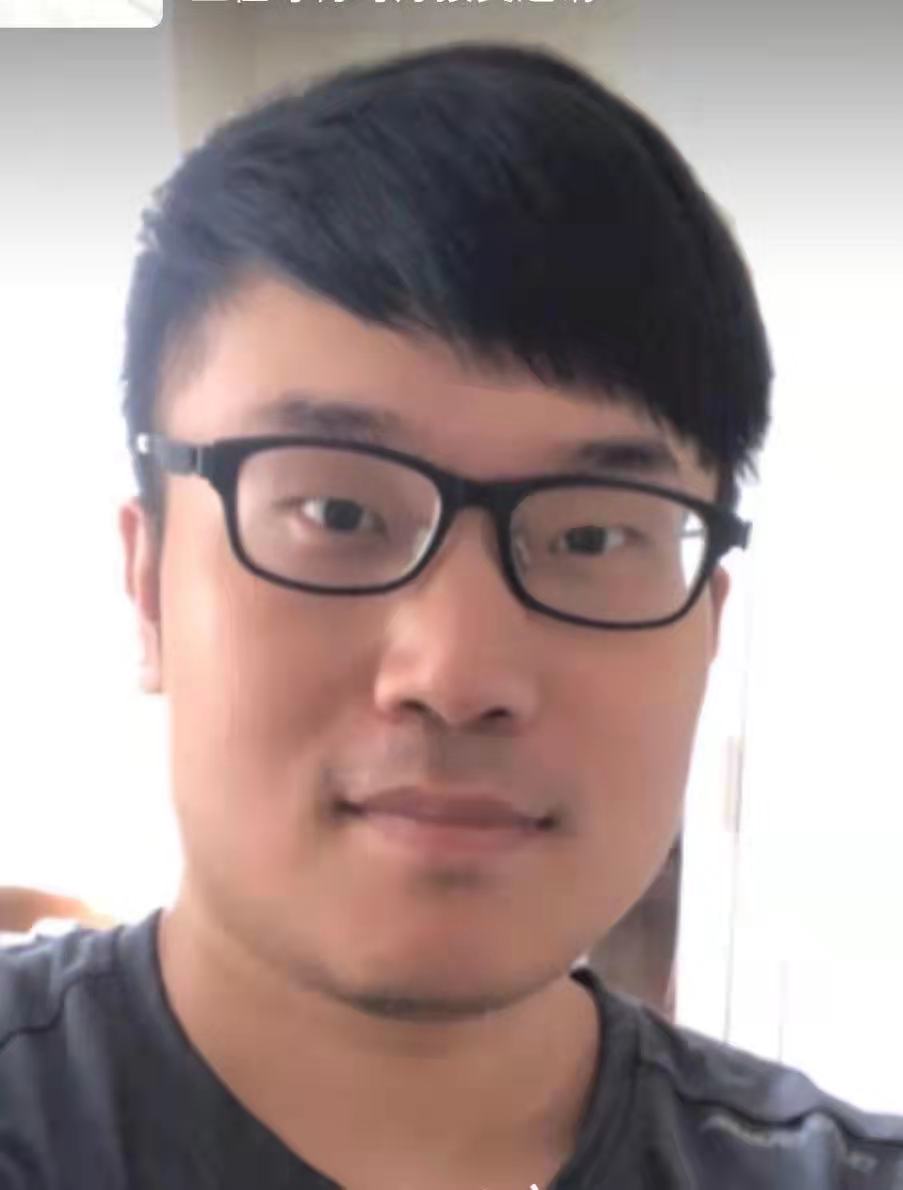}}]{Lei Luo}
 received the Ph.D. degree in pattern recognition and intelligence system from the School of Computer Science and Engineering, Nanjing University of Science and Technology, Nanjing, China. From 2017 to 2020, he was a Post-Doctoral Fellow at the University of Texas at Arlington, Arlington, TX, USA, and the University of Pittsburgh, Pittsburgh, PA, USA. He is currently a data scientist of JD Finance America Corporation, USA. His research interests include pattern recognition and machine learning.
\end{IEEEbiography}

\vspace{11pt}

\begin{IEEEbiography}[{\includegraphics[width=1in,height=1.25in,clip,keepaspectratio]{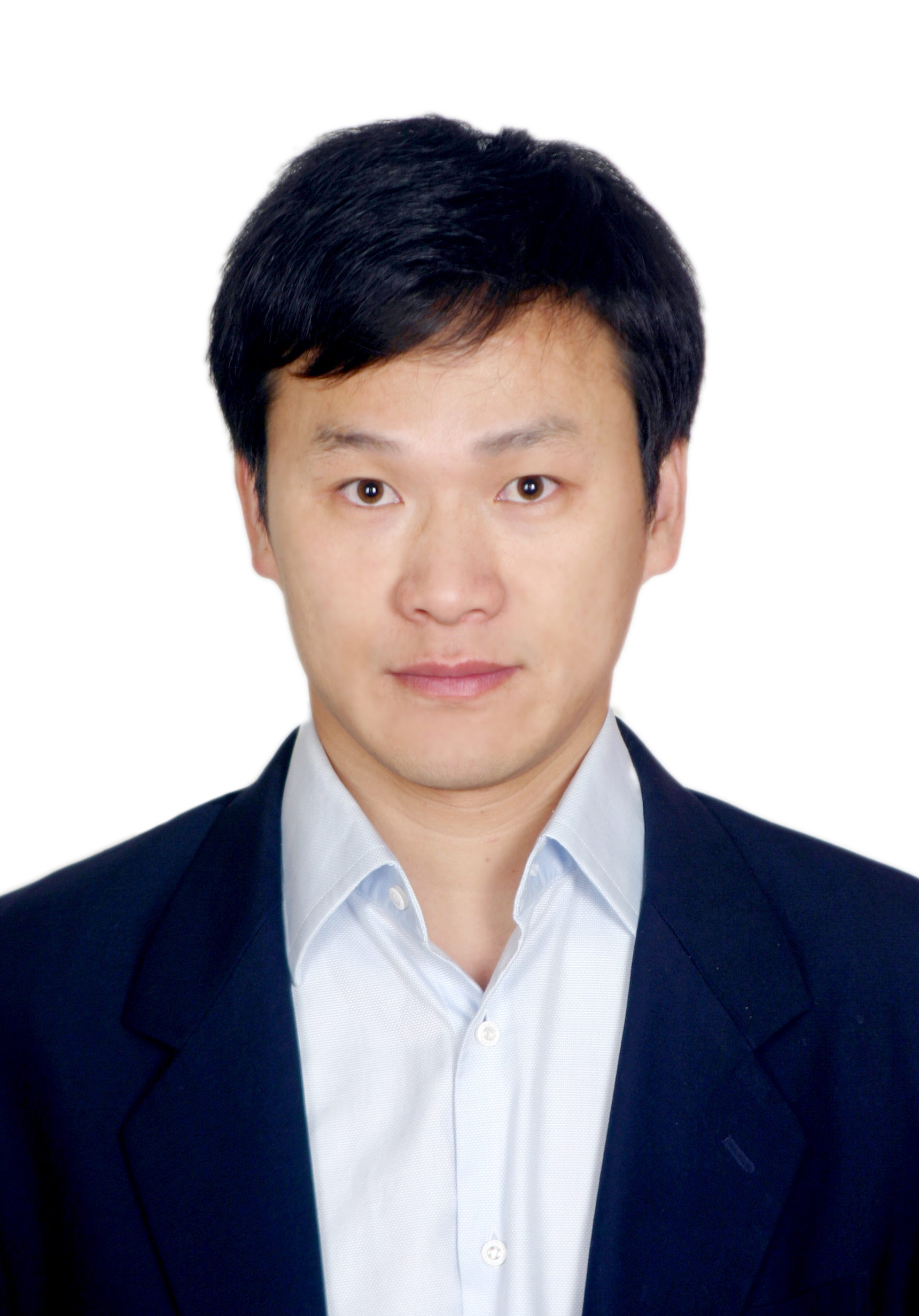}}]{Cheng Deng}
 (Senior Member, IEEE) received the
B.E., M.S., and Ph.D. degrees in signal and information processing from Xidian University, Xi’an,
China, in 2001, 2006, and 2009, respectively.
He is currently a Full Professor with the School
of Electronic Engineering, Xidian University. He has
authored and coauthored more than 100 scientific
articles at top venues, including the IEEE TRANSACTIONS ON NEURAL NETWORKS AND LEARNING
SYSTEMS, TRANSACTIONS ON IMAGE PROCESSING, TRANSACTIONS ON CYBERNETICS, TRANSACTIONS ON MULTIMEDIA, TRANSACTIONS ON SYSTEMS, MAN, AND
CYBERNETICS, the International Conference on Computer Vision, Computer
Vision and Pattern Recognition, the International Conference on Machine
Learning, Neural Information Processing Systems, the International Joint
Conference on Artificial Intelligence, and the Association for Advancement of
Artificial Intelligence. His current research interests include computer vision,
pattern recognition, and information hiding
\end{IEEEbiography}

\vspace{11pt}

\begin{IEEEbiography}[{\includegraphics[width=1in,height=1.25in,clip,keepaspectratio]{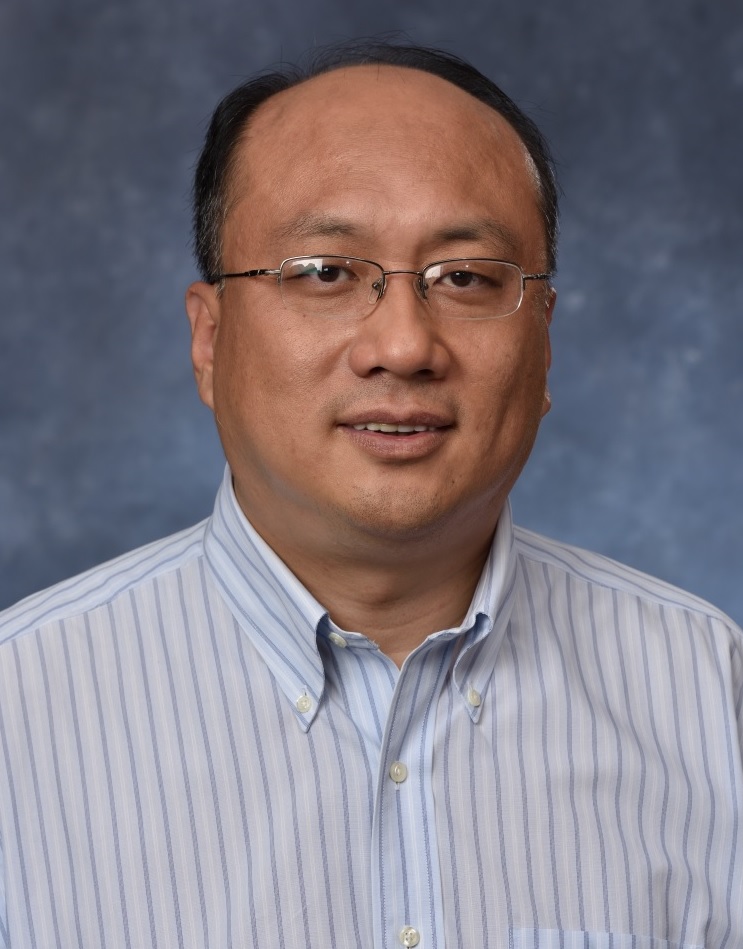}}]{Heng Huang}
received the B.S. and M.S. degrees
from Shanghai Jiao Tong University, Shanghai,
China, in 1997 and 2001, respectively, and the Ph.D.
degree in computer science from the Dartmouth
College, Hanover, NH, USA, in 2006.
He is currently the John A. Jurenko Endowed
Professor in computer engineering with the Department of Electrical and Computer Engineering, University of Pittsburgh, Pittsburgh, PA, USA. He is
also a Consulting Researcher with JD Finance American Corporation, Mountain View, CA, USA. His
research interests include machine learning, data mining, computer vision,
and biomedical data science.
\end{IEEEbiography}

\vspace{11pt}

\end{document}